\documentclass[preprint,journal]{IEEEtran}
\usepackage{xcolor}
\usepackage{amsmath,amsfonts}
\usepackage{algorithmic}            
\usepackage{algorithm}
\usepackage{array}
\usepackage[caption=false,font=normalsize,labelfont=sf,textfont=sf]{subfig}
\usepackage{textcomp}
\usepackage{stfloats}
\usepackage{url}
\usepackage{verbatim}
\usepackage{graphicx}
\usepackage{cite}
\usepackage{float}
\usepackage{xcolor}
\usepackage{multirow}
\definecolor{midnightblue}{rgb}{0.1, 0.1, 0.64}
\newlength{\Oldarrayrulewidth}
\newcommand{\Cline}[2]{%
  \noalign{\global\setlength{\Oldarrayrulewidth}{\arrayrulewidth}}%
  \noalign{\global\setlength{\arrayrulewidth}{#1}}\cline{#2}%
  \noalign{\global\setlength{\arrayrulewidth}{\Oldarrayrulewidth}}}

\begin{document}

\title{Hierarchical Training of Deep Neural Networks \\ Using Early Exiting }

\author{Yamin~Sepehri, Pedram~Pad, Ahmet Caner Y\"{u}z\"{u}g\"{u}ler, \\Pascal~Frossard, and~L.~Andrea~Dunbar
        % <-this % stops a space
\thanks{Yamin Sepehri, Pedram Pad, and L. Andrea Dunbar are with Centre Suisse d'Electronique et de Microtechnique (CSEM), CH 2002 Neuchâtel, Switzerland. Yamin Sepehri is also with the Signal Processing Laboratory (LTS4), École
Polytechnique Fédérale de Lausanne (EPFL), CH-1015 Lausanne, Switzerland.}
\thanks{Ahmet Caner Y\"{u}z\"{u}g\"{u}ler and Pascal Frossard are with the Signal Processing Laboratory (LTS4), École
Polytechnique Fédérale de Lausanne (EPFL), CH-1015 Lausanne, Switzerland}

\footnotesize
\vspace{10pt}
\ This article is accepted to \textit{IEEE Transactions on Neural Networks and Learning Systems}, doi: 10.1109/TNNLS.2024.3396628,  © 2024 IEEE. \url{https://ieeexplore.ieee.org/document/10530344}}% <-this % stops a space
%\markboth{IEEE Transactions on Neural Networks and Learning Systems}%

\maketitle

\begin{abstract}
Deep neural networks provide state-of-the-art accuracy for vision tasks but they require significant resources for training. Thus, they are trained on cloud servers far from the edge devices that acquire the data. This issue increases communication cost, runtime and privacy concerns. In this study, a novel hierarchical training method for deep neural networks is proposed that uses early exits in a divided architecture between edge and cloud workers to reduce the communication cost, training runtime and privacy concerns. The method proposes a brand-new use case for early exits to separate the backward pass of neural networks between the edge and the cloud during the training phase. We address the issues of most available methods that due to the sequential nature of the training phase, cannot train the levels of hierarchy simultaneously or they do it with the cost of compromising privacy. In contrast, our method can use both edge and cloud workers simultaneously, does not share the raw input data with the cloud and does not require communication during the backward pass. Several simulations and on-device experiments for different neural network architectures demonstrate the effectiveness of this method. It is shown that the proposed method reduces the training runtime for VGG-16 and ResNet-18 architectures by $\mathbf{29\%}$ and $\mathbf{61\%}$ in CIFAR-10 classification and by $\mathbf{25\%}$ and $\mathbf{81\%}$ in Tiny ImageNet classification when the communication with the cloud is done over a low bit rate channel. This gain in the runtime is achieved whilst the accuracy drop is negligible. This method is advantageous for online learning of high-accuracy deep neural networks on sensor-holding low-resource devices such as mobile phones or robots as a part of an edge-cloud system, making them more flexible in facing new tasks and classes of data.
\end{abstract}


{\section{Introduction}\label{section:introduction}}

\IEEEPARstart{D}{eep} Neural Networks (DNNs) have shown their effectiveness in different computer vision tasks such as classification \cite{simonyan_very_2015,resnet2015}, object detection \cite{yolo2015,rcnn2014} and body-pose estimation \cite{openpose2018}. These methods outperform the previous classical approaches in all of these areas in terms of accuracy. However, in general, these state-of-the-art DNNs are made of complex structures with numerous layers that are resource-demanding. For example, ResNet-18 is made of 72 layers structured as 18 deep layers and around 11 million trainable parameters~\cite{chandola2021deep}. Implementation of DNNs requires high computational resources to perform many multiplications and accumulations and a high amount of memory to store the vast number of parameters and feature maps. This issue is far more critical in the training phase of DNNs as it is more intense in terms of computations than the inference phase~\cite{adolf_fathom_2016} due to the higher number of FLOPs in the backward pass in comparison to the forward pass and the additional operations needed to update the parameters~\cite{epoch2021backwardforwardFLOPratio}. All these demanding operations are often done in multiple passes in the training phase. Moreover, as the parameters, their updates and the layers' activation maps should be stored, the memory requirements are also higher in the training phase~\cite{celledoni_structure_2020}.

The large resource requirements for training classical DNNs make their implementation often unsuitable for resource-limited edge devices used in IoT systems. The conventional solution to this issue is to offload DNN training to cloud servers, which are abundant in terms of computational resources and memory. 
However, training DNNs on the cloud requires sending the collected input data from the edge to the cloud.This data communication increases the total latency of training, which is crucial when one needs fast online learning and seamless adaptation of models in tasks like human-robot interaction~\cite{qu2022p}.
Additionally, in many tasks, the datasets contain sensitive content such as personal information like identity, gender, etc., which raises privacy concerns if they are sent to the cloud for training a DNN, due to untrusted connections or cloud service providers~\cite{mireshghallah_privacy_2020}. 

To overcome the above problems, different hierarchical training methods have been proposed. The goal of hierarchical training is to train a complex DNN more efficiently by bringing it closer to where the data is acquired by sensors, i.e. the edge worker. As the edge cannot handle the whole training task of the DNN, a part of the training phase is outsourced to another device with substantial resources, i.e., the cloud worker. In other words, it offers a method to efficiently train the DNNs on a heterogeneous hierarchy of workers. Other optional levels may also be added in between, that represent local server workers which are closer to the edge but with lower resources in comparison to the cloud.
For example, in~\cite{eshratifar_jointdnn_2020}, the authors analyzed the different workers as different graph nodes and found the shortest path to decide the schedule of execution on different workers, and in~\cite{liu_hiertrain_2020}, the authors divided the data batches between the different workers with respect to their resources. Although these methods successfully divided the training phase between the workers, they suffered from issues such as high communication cost due to the several data transfers between the workers~\cite{eshratifar_jointdnn_2020}, and privacy concerns as they send directly the raw input data over the network~\cite{liu_hiertrain_2020}.

In this work, we propose a novel hierarchical training framework using the idea of early exiting that provides efficient training of DNNs on edge-cloud systems. The proposed method addresses the issues of training on edge-cloud systems and mitigates the communication cost, reduces the latency, and improves privacy. Our method reduces the latency of training and privacy concerns as it does not send the raw data to the cloud; instead, it shares a set of features only. In order to achieve this goal, we benefit from the idea of early exiting at the edge, which was solely used in full-cloud training to achieve higher accuracy in specific architectures like GoogLeNet \cite{googlenet2014}, or the inference phase of hierarchical systems \cite{teerapittayanon_distributed_2017}. We use early exiting to split the backward pass of training in the DNN architecture parts implemented at the edge and the cloud. It provides the possibility to partially parallelize the training over the edge and the cloud workers, enabling the use of their full potential and reducing the latency. Our early exiting approach does not need to communicate the gradients between the workers in the backward pass. Moreover, it enables the use of non-differentiable operations such as quantization to compress the communicated data in the forward pass. It also has the side benefit of providing robustness against network failures in case where there is an edge-cloud communication blockage during the training phase that causes loss of access to the powerful cloud computation. The edge can resume training separately as its early exit provides the required gradients for executing the backward pass and updating the parameters of the layers at the edge. Additionally, during the inference phase, the early exit can offer a level of classification independently which is often less accurate than the cloud classification, but still acceptable for many applications. Notice that without the early exiting component, when there is a communication blockage, the system is completely out of order. The proposed method further provides a possibility of reduction in the edge power consumption, since there is no need to communicate the backward pass gradients from the cloud back to the edge. 

We perform extensive experiments and compare the performance of our novel model with the baseline of full-cloud training, that is the scenario of communicating the input data directly to the cloud and training the neural network there. We show in on-device experiments that our method can reduce the latency of training in comparison to the baseline while having a negligible amount of accuracy reduction. It improves the training runtime significantly, especially when the communication bandwidth is low. The advantage of our method in terms of computational and memory requirements and communication burden is also shown in our experiments.

Our proposed method is especially useful in online training on resource-constrained devices that acquire their own data from their built-in sensors, such as mobile phones or robots. It allows for training high-accuracy DNNs for the given tasks without the need for high computational and memory resources, or for sharing private raw data with cloud servers. 

The main contributions of this work are summarized as:
\begin{itemize}
    \item We propose a novel approach to train DNNs in hierarchical edge-cloud systems, using early exiting. 
    It results in lower latency, reduced communication cost, improved privacy, and robustness against network failures with respect to a classical full-cloud training framework.
    \item We propose guidelines to efficiently select the partitioning strategy of the DNN between the edge and the cloud based on specific requirements such as runtime, accuracy or memory.
    \item We conduct a performance analysis and show the advantage of the proposed hierarchical training method in terms of memory consumption, computational resource requirements and communication burden. We perform extensive simulations and calculate the latency of different architectures of DNNs when they are trained with our edge-cloud framework and show their superiority over the full-cloud training while having a negligible accuracy drop.
    \item We implement the system and perform on-device experiments to show the runtime improvements in an experimental edge-cloud setup to validate the proposed idea.
    \item We demonstrate the effectiveness of different parts of the proposed hierarchical training method in the improvement of accuracy, communication requirements and training runtime of the DNN in a comprehensive ablation study.
\end{itemize}

The manuscript is structured as follows: the related works are described in Section~\ref{section:relatedworks}.
After that, the proposed hierarchical training approach is elaborated in Section~\ref{section:HierTrainEarly}. Section~\ref{section:Exp} is dedicated to the experiments and their results. Section~\ref{section:ablation_study} presents ablation studies illustrating the efficacy of each component in the proposed method. The study is concluded in Section~\ref{section:conclusion}.

{\section{Related Works}\label{section:relatedworks}}
In this section, we describe the related works and divide them into three groups: hierarchical training, early exiting, and hierarchical inference. 
\vspace{5pt}
{\subsection{Hierarchical Training}\label{section:relhiertrain}}

The idea of training a DNN on a system of hierarchical workers with different available resources has recently received attention from researchers. In these works, the authors try to train a DNN directly on an edge-cloud system. Eshratifar~et~al.~\cite{eshratifar_jointdnn_2020} proposed the idea of JointDNN that defines the blocks of a DNN that are executed on each worker as graph nodes and proposed a method of DNN division by solving the shortest path problem in this graph. Although their method is able to improve the runtime and energy consumption of the training phase, it cannot benefit all levels of the hierarchy at the same time due to the sequential nature of the training phase. Additionally, the communication cost is still high between the mobile and the cloud levels as their proposed method sometimes has more communication stages than the two usual ones (for the forward pass and the backward pass).

Liu et al. \cite{liu_hiertrain_2020} proposed the idea of HierTrain that benefits from hybrid parallelism to use the capacity of the different workers. They proposed a method that finds an optimized division point of the model between the workers and is also able to divide the input data in terms of batches with different sizes between the different levels of the hierarchy. Their method is able to reduce the latency of execution with respect to the full-cloud framework; however, it compromises the privacy of the users. The hybrid parallelism method sends a significant amount of raw input data samples directly to the cloud. Sending this raw data to the higher level workers also increases the communication cost of their method, in addition to the privacy issues. Moreover, this method is vulnerable when network connection failures happen.

In contrast to these works, our goal is to propose a hierarchical training method that has a low communication cost and respects the privacy of users by not communicating the raw data. This method should be able to exploit the computational potentials of the edge and the cloud at the same time.
\vspace{2pt}
{\subsection{Early Exiting}\label{section:relearlyexit}}

Early exiting is a method in DNNs that performs the decision-making in earlier layers in addition to the last layer. In non-hierarchical systems, it has been used in the training of DNNs in architectures such as GoogLeNet \cite{googlenet2014} to achieve better accuracy of inference. In these methods, the early exit is used during the training phase and is removed in the inference phase. In hierarchical systems, there are methods such as \cite{teerapittayanon_distributed_2017} and \cite{wang_adda_2019} that use early exiting to reduce the communication burden of DNNs' inference. In these models, the samples that have high confidence levels in the early exit are classified at the edge while the others are sent to the cloud for making the decision. 

In contrast, in our work, the focus is on the benefit of early exits to improve the training phase of hierarchical edge-cloud systems. It reduces the latency of the training phase by parallelizing it on the edge and the cloud workers and provides robustness against network failures.
\vspace{5pt}
{\subsection{Hierarchical Inference}\label{section:relhierinference}}

To be complete on the related studies, we also discuss the hierarchical inference methods. In this group of works, the authors proposed a hierarchical framework for the inference phase of a DNN. Teerapittayanon et al.~\cite{teerapittayanon_distributed_2017} proposed a distributed computing hierarchy that is made of three levels of cloud, fog and edge where they can execute the inference phase of a DNN when it is separated between them. They benefit from early exiting to reduce the required communication cost between the different levels. After that, Wang et al.~\cite{wang_adda_2019} proposed adaptive distributed acceleration of DNNs where they proposed a method to select the best position to divide neural networks between the two workers in the inference phase. Recently, Xue et al. \cite{xue_ddpqn_2022} proposed a more advanced algorithm that, instead of using iterative approaches to find the best position of separation on hierarchical systems, benefits from the decision-making ability of reinforcement learning to perform the offloading strategy, in systems with complex conditions. Another idea that has been used to reduce the communication cost, relates to using methods of lossy and lossless compression on the communicated data \cite{li_jalad_2018}.
\begin{figure*}[!t]
\centering
\includegraphics[scale=0.38]{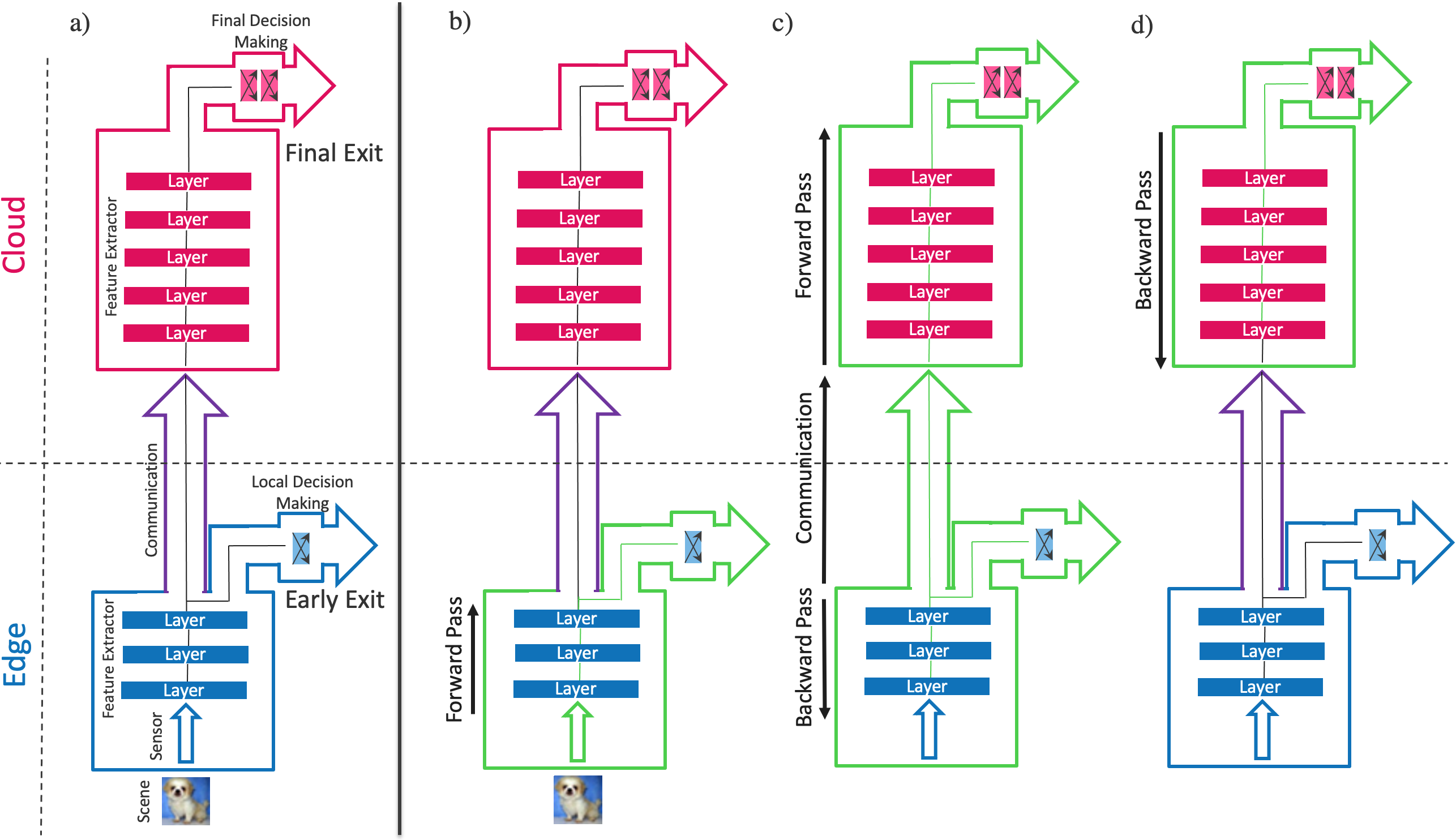}
\caption{a) A schematic view of the different parts in the proposed hierarchical execution framework. b) In the first step of training, the forward pass is done at the edge feature extractor to generate the feature set that is sent to the cloud. A local decision
maker also takes the output of this feature extractor enabling
an early exit that is later used for the backward pass at the edge. c) The feature set is communicated to the cloud and a more computationally intensive feature extraction together with a final decision-making are done there. Simultaneously, the backward pass of the edge is done to train the edge parameters. d) The backward pass of the cloud is done to update the cloud parameters. The green borders indicate running processes at each step.}
\label{fig:hier_model}
\end{figure*}

Although these methods are effective in executing the inference phase on the edge-cloud frameworks, they do not consider the problem of training, which is more complex.

\vspace{5pt}
{\section{Hierarchical Training with Early Exiting}\label{section:HierTrainEarly}}

{\subsection{Proposed Hierarchical Training Framework}\label{section:framework}}
As mentioned in Section~\ref{section:relhiertrain}, the idea of training DNNs on a hierarchy of multiple workers is a new area of research that is not yet fully explored. The previous works generally do not exploit the potential of all levels and have high communication costs and privacy concerns. Our method addresses these issues by using an early exiting scheme, that enables a level of parallelism between the edge and the cloud in the different steps of the training phase.

More specifically, we divide a conventional DNN architecture between an edge worker and a cloud worker at one of the middle layers, assuming the cloud has higher computational resources in comparison to the edge device, which is often the case (Figure~\ref{fig:hier_model}-a). In the first step of training (shown in Figure~\ref{fig:hier_model}-b), the data acquired by the sensor at the edge passes through the layers of the neural network that are located at the edge worker, and the forward pass of the edge is executed. This neural network has two main building blocks: a feature extractor and a local decision-maker. The edge feature extractor's output is transmitted to the cloud to be processed by the remaining layers of the DNN architecture. The local decision-maker also takes the output of this feature extractor enabling an early exit. The edge loss that is generated in this early exit later allows the neural network parameters at the edge to be updated during the backward pass.

 In the second step (shown in Figure~\ref{fig:hier_model}-c), the feature map, which is generated by the feature extractor at the edge, is communicated to the cloud server. The forward pass of the cloud layers of the DNN is completed, allowing a more computationally intensive feature extraction to take place. Then, the final output generates the cloud loss using the targets. At this time, the backward pass of the edge layer is also done at the edge level for training the parameters of the edge layers. In other words, these two tasks are done in parallel. In the third step (shown in Figure~\ref{fig:hier_model}-d), the backward pass of the cloud layers is done to update the parameters of these layers.

In addition to exploiting the potential of both workers simultaneously, an important benefit of this approach is that there is no need to perform any communication during the backward pass, in contrast to previous works. (e.g., \cite{liu_hiertrain_2020}). The backward pass of each worker is done independently, resulting in a significant reduction in the communication cost and the total runtime. Indeed, the edge worker only transmits data and does not need to have the receiving ability. Thus, a less complex communicator device is required at the edge which is favorable in practice. 

Another benefit of this method is that, as no backward pass happens in the position between the edge and the cloud, non-differentiable functions can be applied to the feature map to further compress it before the communication. For example, the activations can be quantized before communicating to the cloud server, reducing the communication burden. These lossy compression methods should however be implemented while considering their possible penalty on the total accuracy.

The training phase is done for a number of iterations selected by the user. After finishing it, the inference phase is done on the proposed edge-cloud framework. The steps of the inference phase are simply the forward pass at the edge, the communication of the feature map to the cloud, the forward pass of the cloud and the decision-making using the final exit at the cloud. It is worth mentioning that the early exit at the edge device can be also used as a local decision-maker in case of network failures.
\vspace{4pt}
{\subsection{Runtime Analysis}\label{section:runtime_analysis}}

One of our main goals is to improve the overall runtime of the training phase in our proposed hierarchical training method in comparison to the full-cloud training. This target can be achieved by splitting a neural network at a suitable point between the edge and the cloud, based on the performance of the devices, the selected communication protocol and the selected DNN architecture. However, the training runtime cannot be estimated easily without physically implementing it on the devices. The reason is the different choices for the devices and the different internal structures and delays that result in different overall runtimes. To tackle this issue, we propose a method to simplify this procedure and estimate the runtime by just performing one epoch of forward pass at the selected edge and the cloud. 
We use this forward pass runtime to estimate the runtime of the backward pass. We also calculate the runtime of the update phase and communication phase and propose a method to sum them up to compute the estimated training runtime.

We propose Equation~\ref{eq:tot_hier} to estimate the training runtime of our proposed hierarchical training method.
\begin{equation}\label{eq:tot_hier}
\begin{split}
    T^\text{ hierarchical}_\text{total} =&\ T^\text{ edge}_\text{comp,forw} + \max \Big(  \big(T^\text{ hierarchical}_\text{comm}+ \\& T^\text{ cloud}_\text{comp,forw} + T^\text{ cloud}_\text{comp,backw} \big),T^\text{ edge}_\text{comp,backw} \Big)
\end{split}
\end{equation}
where $T^\text{ hierarchical}_\text{total}$ is the total runtime, $T^\text{ edge}_\text{comp,forw}$ and $T^\text{ edge}_\text{comp,backw}$ are the runtimes of the forward pass and the backward pass of the edge part of the DNN architecture,  $T^\text{ cloud}_\text{comp,forw}$ and $T^\text{ cloud}_\text{comp,backw}$ are the runtimes of the forward pass and the backward pass of the cloud part of the DNN architecture and $T^\text{ hierarchical}_\text{comm}$ shows the duration of communication between the edge and the cloud. The $\max$ function reflects that, in our proposed hierarchical training method, the backward pass at the edge is executed at the same time as the communication and execution on the cloud. 

Only one forward pass is done on the connected edge and cloud to measure the values of $T^\text{ edge}_\text{comp,forw}$ and $T^\text{ cloud}_\text{comp,forw}$. Afterwards, we calculate the backward pass runtime terms in Equation~(\ref{eq:tot_hier}) as 

\begin{equation}
    T^\text{ edge}_\text{comp,backw} = \alpha \  T^\text{ edge}_\text{comp,forw} +  \beta \sum^{e}_{i=1} \frac{P_{i}}{S^\text{ edge}},
\label{eq:back_edge}
\end{equation}
\begin{equation}
    T^\text{ cloud}_\text{comp,\text{backw}} = \alpha \  T^\text{ cloud}_\text{comp,forw} +  \beta \sum^{c}_{i=e+1} \frac{P_{i}}{S^\text{ cloud}} .
\label{eq:back_cloud}
\end{equation}
In Equation (\ref{eq:back_edge}) and (\ref{eq:back_cloud}), the first term shows the time which is needed to perform calculations of the backward pass. We calculate this by multiplying the backward-to-forward ratio $\alpha$ by the measured forward pass runtime. In general, the scaler $\alpha$ is a value that for most DNNs with convolution layers and large batch sizes is close to $2$~\cite{epoch2021backwardforwardFLOPratio}. The second term shows the update phase duration and the summations indicate all the layers that are implemented at the edge or the cloud, where $e$ represents the last layer that is executed on the edge and $c$ is the total number of layers in the DNN. $P_i$ represents the number of parameters in layer $i$ that should be updated. $S$ indicates the theoretical performance (computation speed) of each of the edge and the cloud devices. The scalar $\beta$ shows the number of FLOPs that are needed to update every parameter of the DNN using the selected optimizer. As an example, stochastic gradient decent requires $2$ FLOPs per parameter and Adam optimizer~\cite{kingma2014adam} requires $18$ FLOPs per parameter~\cite{epoch2021backwardforwardFLOPratio}, that results in $\beta = 2 \text{ or } 18$ for these cases.

The communication duration in Equation~(\ref{eq:tot_hier}) can be calculated as 
\begin{equation}\label{eq:comm}
    T^\text{ hierarchical}_\text{comm} = \frac{D_\text{comm}}{S_\text{network}}
\end{equation}
where $D_\text{comm}$ is the size of data that is communicated. For our proposed method of hierarchical training, it is the size of the feature map that is communicated. $S_\text{network}$ represents the bandwidth of the selected communication network.

In this study, we compare our work with the full-cloud training method as the baseline. In this case, the edge device just captures the inputs and transmits them to the cloud and the whole training procedure for all layers of the DNN is done on the cloud. The runtime of the full-cloud method is simply calculated as
\begin{equation}\label{eq:tot_full}
\begin{split}
    T^\text{ fullcloud}_\text{total} =&\ T^\text{ fullcloud}_\text{comm}\ + T^\text{ cloud}_\text{comp,forw} + T^\text{ cloud}_\text{comp,backw} \ . 
\end{split}
\end{equation}

The backward pass term of Equation~(\ref{eq:tot_full}) can be similarly calculated with one forward pass runtime measurement by exploiting Equation~(\ref{eq:back_cloud}). The communication term of Equation~(\ref{eq:tot_full}) can also be calculated with Equation~(\ref{eq:comm}). The only difference is the $D_\text{comm}$ that is now the size of the raw input data that is communicated to the cloud.

\vspace{5pt}
{\subsection{Separation Point Selection}\label{section:separation_selection}}
In the proposed hierarchical method, selecting the position of the separation point of the DNN between the edge and the cloud is important as it can affect the computational and memory burden on the edge and the cloud, the total runtime, and the accuracy. In this section, we propose an algorithm to select the edge-cloud separation position based on user requirements. As the full training of DNNs is computationally heavy, it is highly demanding to try all possible splitting points. Hence, the proposed algorithm smartly confines the splitting point candidates in order to lower the number of full-training trials. The user requirements may contain a maximum possible number of parameters at the edge (memory), a maximum duration of training (runtime) and a minimum precision (accuracy).

We propose Algorithm~\ref{alg:acc_runtime} for this purpose. The algorithm benefits from Equations~(\ref{eq:tot_hier})-(\ref{eq:comm}) to reduce the number of full-training iterations needed in different separation points and to finally find a good position of separation. It finds the desired separation point for the given hardware specifications.

\begin{algorithm} [t]
 \caption{Selecting the separation point of a DNN between edge and cloud based on specific runtime and accuracy criteria}
 \begin{algorithmic}[1]
 \renewcommand{\algorithmicrequire}{\textbf{Input:}}
 \renewcommand{\algorithmicensure}{\textbf{Output:}}
 \REQUIRE DNN, Available Edge Memory, Speed of Computation (edge and cloud) and Communication, Accepted Rutnime, Accepted Accuracy
 \ENSURE Position of Separation $P$
 \\ \textit{Initialization} :
  \STATE $L^i |i = 1 : N$ : layers/blocks in the DNN
  \STATE $S_1,S_2,S_3 \leftarrow  \emptyset$
  \\ \textit{Loop 1} :
  \FOR {$L^i |i = 1 : N$}
  \STATE \textbf{Measure} $M^{ i}_\text{edge}$ by counting parameters at the edge
  \IF{{$M^{ i}_\text{edge} < \text{Available Edge Memory}$}}
  \STATE $S_1.$\textbf{append }$(i)$
  \ENDIF
  \ENDFOR
  \\ \textit{Loop 2} :
  \FOR {$j \in  S_1$}
  \STATE \textbf{Calculate} $T^{ j}_\text{hierarchical,calc}$ by Eq. (\ref{eq:tot_hier})-(\ref{eq:comm})
  \IF{{$T^j_\text{hierarchical,calc} < \text{Accepted Runtime}$}}
  \STATE $S_2.$\textbf{append }$(j)$
  \ENDIF
  \ENDFOR
  \\ \textit{Loop 3} :
  \FOR {$ l \in  S_2$}
  \STATE \textbf{Train} DNN separated at $l$ for 1 epoch
  \STATE \textbf{Measure} $T^{ l}_\text{hierarchical,exp}$
  \IF{$T^l_\text{hierarchical,exp} < \text{Accepted Runtime}$}
  \STATE $S_3.$\textbf{append }$(l)$
  \ENDIF
  \ENDFOR
  \STATE  $S_3 \leftarrow \textbf{sort } 
  \{S_3| \text{in descending order on } m,m \in S_3$\}
  \\ \textit{Loop 4} :
  \FOR {$ n \in  S_3$}
  \STATE \textbf{Train} DNN for the rest of epochs
  \IF{$Acc^n_\text{hierarchical,exp} > \text{Accepted Accuracy}$}
  \STATE $P \leftarrow n$
  \STATE \textbf{break}
  \ENDIF
  \ENDFOR

 \RETURN{$P$}
 
 \end{algorithmic}
 \label{alg:acc_runtime}
 \end{algorithm}

\begin{figure*}[!h]
\centering
\includegraphics[scale=0.29]{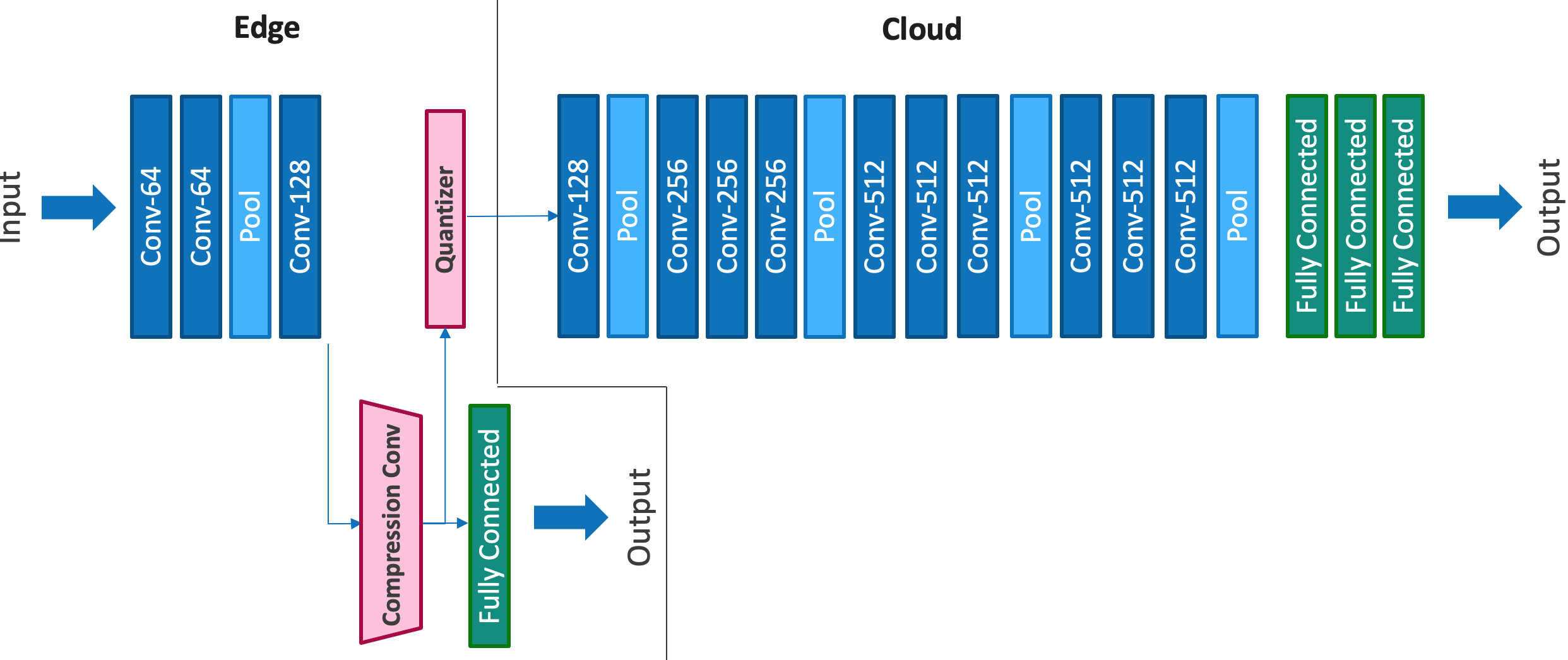}
\caption{The structure of the proposed hierarchical training method applied on VGG-16 architecture \cite{simonyan_very_2015}. The position of separation can be moved along the different layers of the architecture.}
\label{fig:vgg_arch}
\end{figure*}
\begin{figure*}[!h]
\centering
\includegraphics[scale=0.35]{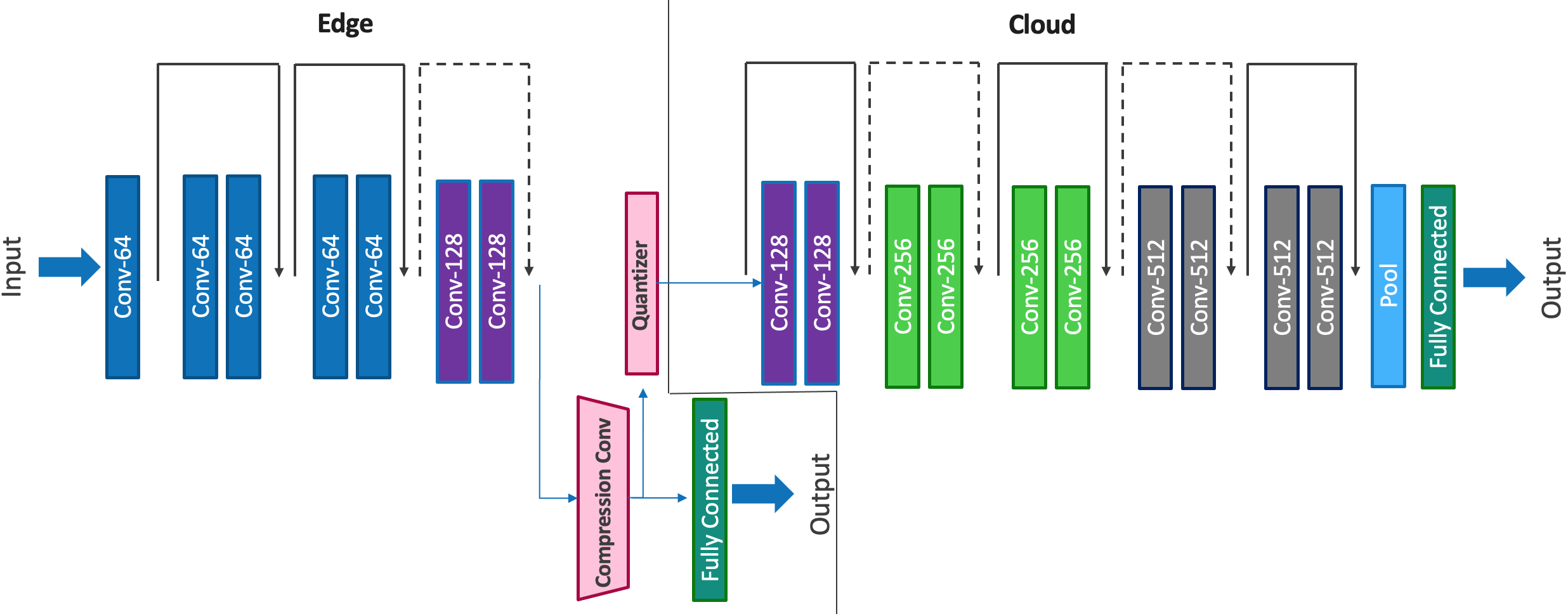}
\caption{The structure of the proposed hierarchical training method applied on ResNet-18 architecture \cite{resnet2015}. The position of separation can be moved along the different residual blocks of the architecture.}
\label{fig:resnet_arch}
\end{figure*}

The algorithm inputs are the DNN architecture, the memory of the edge device, the computational power of the edge and cloud devices, the communication speed between the edge and the cloud and the accepted runtime and accuracy selected by the user. In Loop~1, we consider the memory requirement. We simply measure the number of parameters when the architecture is separated on different points. We add the separation points that satisfy the user's edge memory criterion to set $S_1$.

In Loop~2, we compute the runtime based on Equations (\ref{eq:tot_hier})-(\ref{eq:comm}) for each possible separation point in $S_1$. If for a separation point, the calculated training time is less than the acceptable criterion ($T^j_\text{hierarchical,calc} < Accepted\ Runtime$), it is added to set $S_2$. As we mentioned in Section \ref{section:runtime_analysis}, it is required to measure the time of just one forward pass before doing these calculations. 
 
 In Loop~3, the network is trained for one epoch for these separation points at Set $S_2$ to measure the experimental runtime $T^l_\text{hierarchical,calc}$. If the experimental runtime is acceptable for a separation point, it is added to set $S_3$. Afterwards, $S_3$ is sorted based on the separation points, from the deepest to the earliest. The reason for this strategy is that we observe that the deeper separation points provide better accuracy. It will be shown in the experiments in the next sections.
 
 In Loop~4, the network is trained for the rest of the epochs to calculate the accuracy of the separation points selected from the sorted $S_3$. If the accuracy is higher than the accepted value, the separation point is reported and the procedure is finished; otherwise, the next separation point from the set is selected and the same procedure is repeated. Notice that, in order to measure the accuracy, there is no way but to train the network completely; however, our method requires a low effort to achieve this goal by doing it on a carefully confined and sorted set.

 In this algorithm, the full training just exists in Loop 4, after we limit the size of possible candidate separation points in Loop~1 by a measurement of parameters' size and in Loop~2 by a calculation of estimated runtime. Moreover, in Loop 3, we again reduce the size of possible separation points by performing only $1$ epoch of training. This often results in few splitting points left to be fully trained in Loop~4. Loop~4 also might not be done completely since we sorted the set of possible separation points for the loop in a way that there is a higher chance of achieving the required accuracy in the first iterations.

 \begin{figure*}[h]
\centering
\includegraphics[scale=0.4]{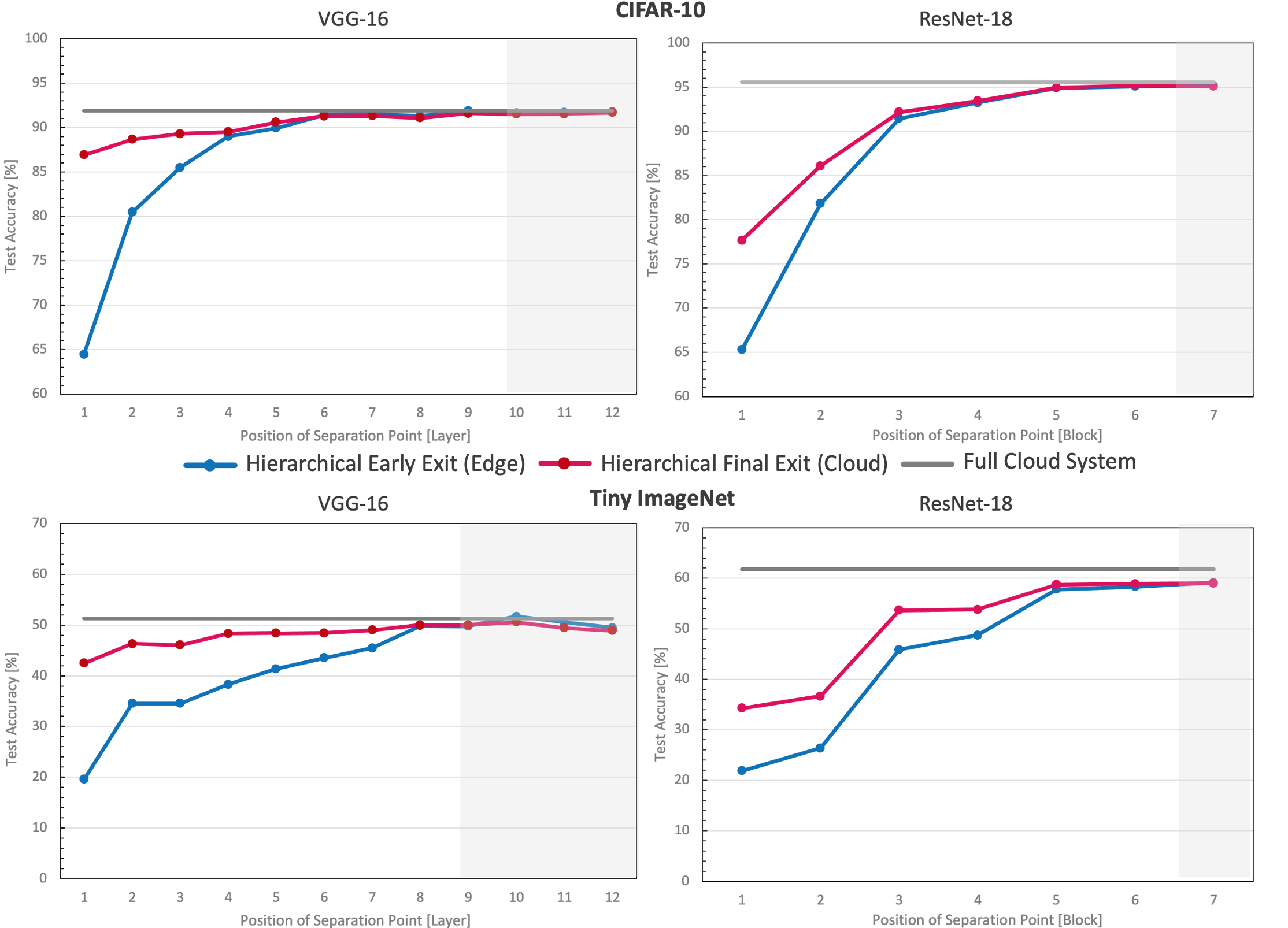}
\caption{The test accuracy of experiments on CIFAR-10 and Tiny ImageNet datasets for hierarchical training when separated at different points along the architectures compared to the full-cloud training accuracy. The left figures show the results for VGG-16 and the right figures show the results for ResNet-18. In addition to the main accuracy, the early exit test accuracy is presented as a side benefit of the proposed method. Although this accuracy is lower in comparison to the final exit, it shows how the edge handles the classification problem independently when there is a possible communication failure. The deepest separation points on the right side (the gray areas) are not practically desirable in the proposed hierarchical training method due to the high memory pressure at the edge and are shown for the sake of completeness (see Figure \ref{fig:memory}).}
\label{fig:accuracy}
\end{figure*}
\begin{figure*}[h]
\centering
\includegraphics[scale=0.39]{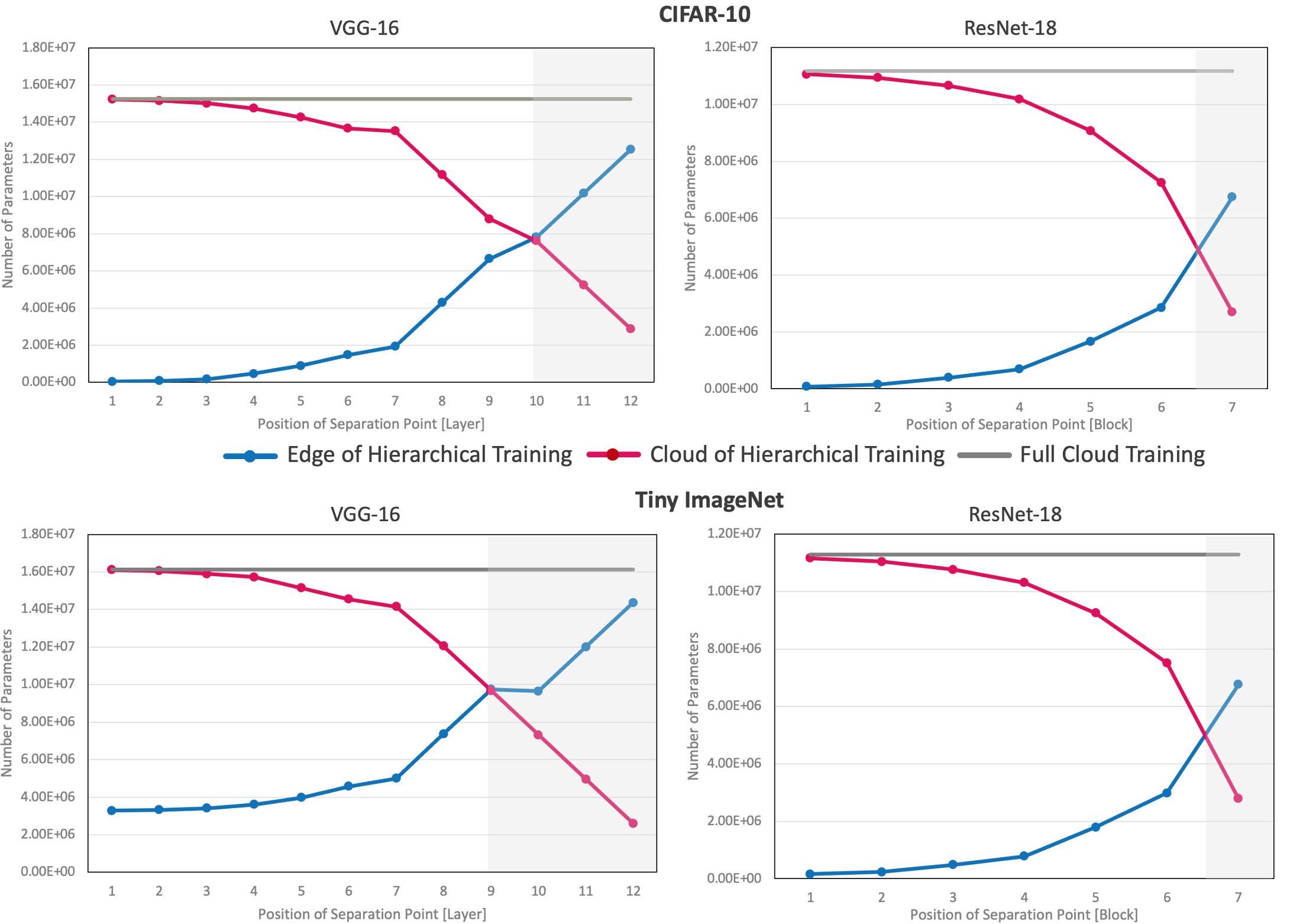}
\caption{The number of parameters of the deep neural network at the edge side and the cloud side in the hierarchical system when it is separated at different points along the architecture in comparison to the number of parameters in the full-cloud system for CIFAR-10 and Tiny ImageNet experiments. The left figures show the results for VGG-16 and the right figures show the results for ResNet-18. Notice the low number of parameters at the edge for most of the separation points in the proposed hierarchical training method that is desirable due to the possible memory constraints. The deepest separation points on the right side (the gray areas) show where the number of parameters at the edge in the proposed hierarchical training method rises above the cloud. These points are not practically desirable due to the high memory pressure at the edge and are shown for the sake of completeness.}
\label{fig:memory}
\end{figure*}
 
 It is worth mentioning that Algorithm~\ref{alg:acc_runtime} can be modified easily for simpler scenarios. For example, when the user has no accuracy requirements, the only difference is that Loop~4 should be removed and all the separation points of Set $S_3$ are acceptable, or when the user has no memory criterion, Loop~1 should be omitted.

In the next section, we perform experiments to validate the benefits of the proposed hierarchical training method. 

{\section{Experiments}\label{section:Exp}}

In this section, we elaborate on the hierarchical training framework for different DNN architectures, describe the experiments performed using the proposed method and report the results. These particular DNN architectures are selected as they are widely used in vision tasks, and they are relatively intensive in terms of resource requirements which makes them challenging to run on low-resource devices.
\vspace{2pt}
{\subsection{Hierarchical Models}\label{section:hier_models}}
Figure \ref{fig:vgg_arch} shows VGG-16 \cite{simonyan_very_2015} architecture implemented in a hierarchical fashion. In this figure, the network is divided between the edge and the cloud after the third convolution layer. This position of separation can however be moved along all layers based on the user's requirements. 

\begin{figure*}[h]
\centering
\includegraphics[scale=0.41]{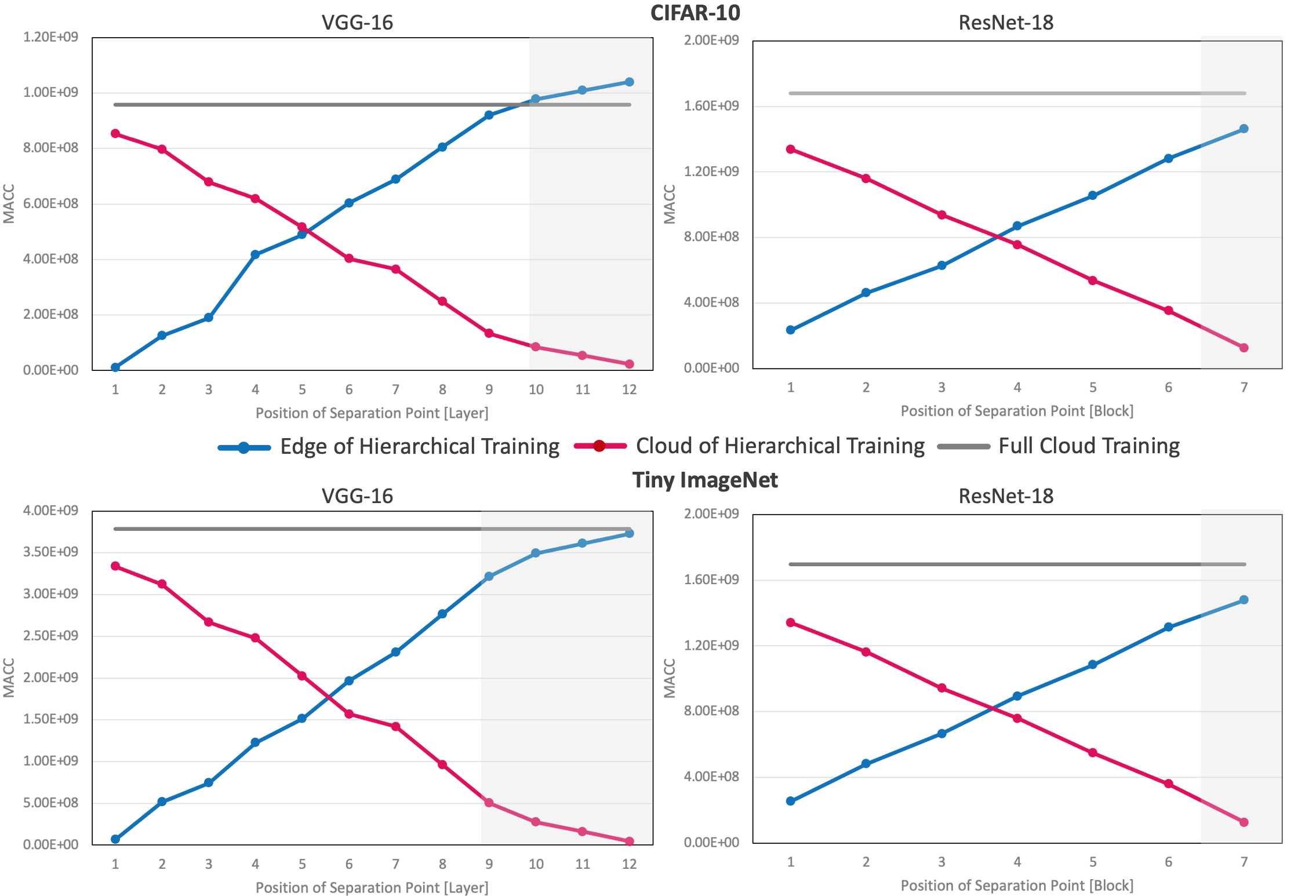}
\caption{The computational cost of the deep neural network in terms of MACC at the edge side and the cloud side during the hierarchical training when it is separated at different points along the architecture in comparison to the full-cloud training in CIFAR-10 and Tiny ImageNet experiments. The left figures show the results for VGG-16 and the right figures show the results for ResNet-18. Notice the low computational burden at the edge for many separation points in hierarchical training that makes them desirable since the computational power is often constrained. The deepest separation points on the right side (the gray areas) are not practically desirable in the proposed hierarchical training method due to the high memory pressure at the edge and are shown for the sake of completeness (see Figure \ref{fig:memory}).}
\label{fig:compute}
\end{figure*}
\begin{figure*}
\centering
\includegraphics[scale=0.41]{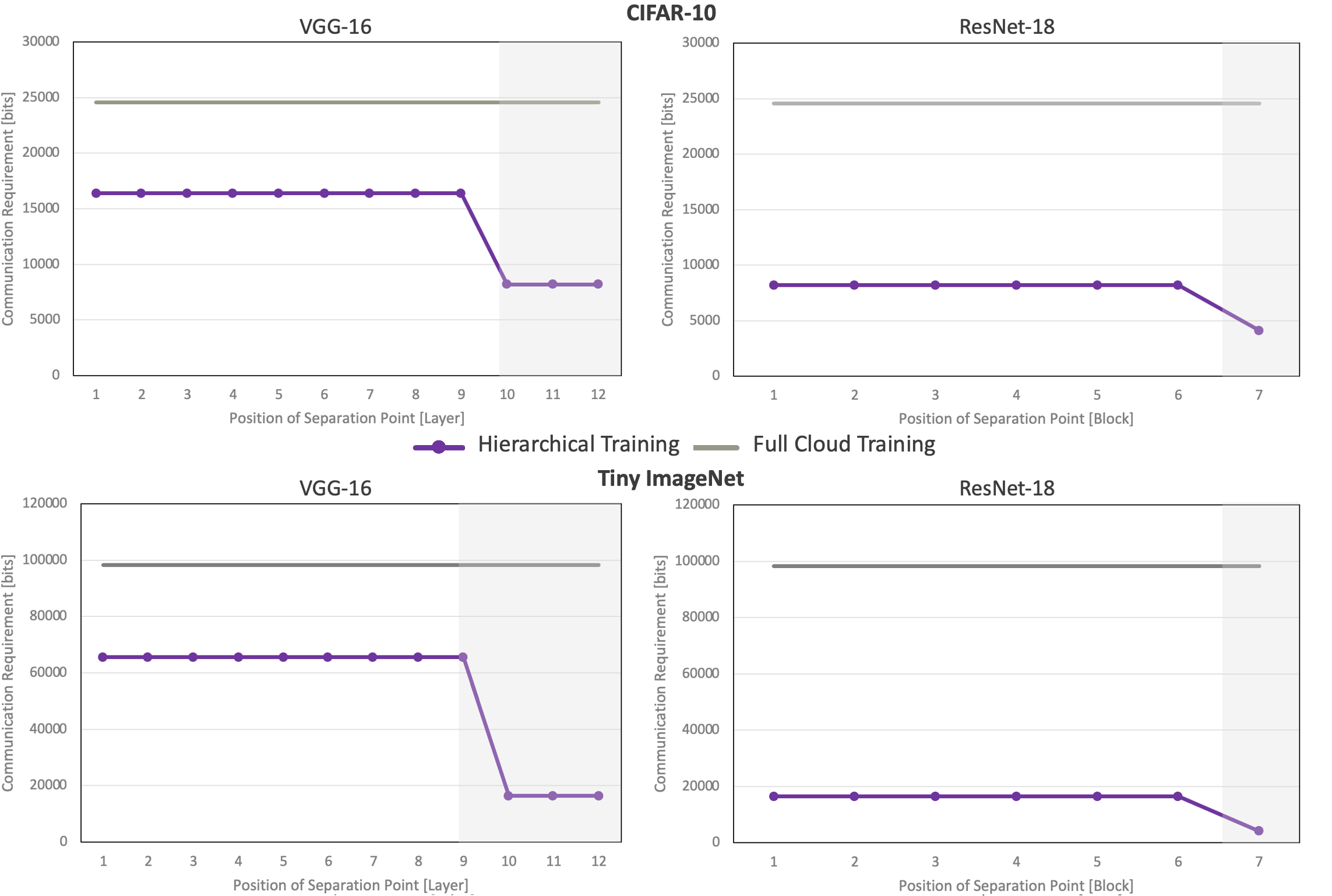}
\caption{The required communication burden in terms of bits by the proposed hierarchical training method when it is separated at different points along the architecture compared to the full-cloud training baseline that directly sends the original inputs to the cloud in CIFAR-10 and Tiny ImageNet experiments. The left figures show the results for VGG-16 and the right figures show the results for ResNet-18. The proposed hierarchical training approach has a lower communication burden in comparison to the full-cloud training that communicates the original inputs. The deepest separation points on the right side (the gray areas) are not practically desirable in the proposed hierarchical training method due to the high memory pressure at the edge and are shown for the sake of completeness (see Figure \ref{fig:memory}).}
\label{fig:communication}
\end{figure*}
 
The output of early exit provides the loss for the backward pass at the edge and the final output for the cloud provides the loss for the backpropagation at the cloud. The important point here is that most conventional DNN architectures such as VGG are not designed to be hierarchically executed. As a result, after most of their layers, the size of the feature map is large, even in comparison to the original input image which makes them expensive to communicate over the network. This increase in the communication cost is not desirable in the hierarchical execution of DNNs. In order to address this issue, a compression convolution layer and a quantizer are used at the division point which reduce the size of the feature map before sending it to the cloud. 
In our experiments, a $4$-bit quantization is used. This bit width is the maximum value that provides a lower communication burden for all the possible separation points of our DNNs, in comparison to the full-cloud training.

Since the range of the feature map may change significantly during the training phase \cite{koster2017flexpoint}, for every quantized batch that is sent to the cloud, one full precision scale value is also communicated \cite{cambier_shifted_2020}. The scale value is calculated as the difference between the maximum value and the minimum value in that batch of data divided by the maximum quantization level ($2^4-1 = 15$ in our case). The minimum value is $0$ since ReLU activation functions are used. All the members of the batch are divided by this scale and then quantized before communicating to the cloud. The scale is also communicated to the cloud and is multiplied by the values of the batch there again. This single scale for every batch has a negligible impact on the communication cost, but a significant role in covering the data range to keep the total accuracy high during the different training iterations. Finally, in order to avoid a high memory burden for the edge worker, only one fully connected layer is used in the early exit.

The same form of hierarchical implementation is applied on ResNet-18 \cite{resnet2015} architecture. Figure \ref{fig:resnet_arch} shows the proposed structure. There is again the compression convolution and the quantizer, and the edge-cloud separation can be done after every residual block in this architecture. 

We note that the two DNN architectures of VGG-16 and ResNet-18 are analyzed in this study as they are relatively large models that cannot be easily trained on low-resource devices. The results can represent a large group of convolutional neural networks and residual neural networks that are widely used in vision tasks.

\vspace{5pt}
{\subsection{Performance Analysis}\label{section:perf_analysis}}
In this part, the performance analysis setup is elaborated. We analyze the important performance factors of accuracy, memory footprint, computational burden, communication rate and runtime in our proposed hierarchical training framework and compare the results with the baseline of full-cloud training. The experiments are done on CIFAR-10 dataset with 10 classes of $32\times32$ images \cite{Krizhevsky_2009_learning} and Tiny ImageNet dataset with $200$ classes of $64\times64$ images \cite{le2015tiny, deng2009imagenet}.
The VGG-16 architecture is trained for $100$ epochs with an Adam optimizer~\cite{kingma2014adam} with a learning rate of $2\times10^{-4}$ without early stopping. The number of output channels of the compression convolution varies between $4$ and $512$ based on the position of the separation point. The ResNet-18 architecture is trained for $200$ epochs without early stopping with a stochastic gradient descent optimizer and an initial learning rate of $0.1$. This learning rate undergoes reduction at each epoch using a cosine annealing scheduler~\cite{loshchilov_sgdr_2017}. The number of channels in the output of the compression convolution varies between $4$ and $64$ based on the position of the separation point. For both architectures, we used random crop, random horizontal flip and normalization preprocessing methods before feeding the data samples to the model in the training phase. In the random crop, the images are firstly zero-padded by $4$ pixels on each side and then randomly cropped by a size of $32\times32$ for CIFAR-10 and $64\times64$ for Tiny ImageNet. For both models and both datasets, the training phase batch size is $64$ and the testing phase batch size is $50$. The early exit and the final exit in both architectures use similar cross-entropy loss. To keep the comparisons fair, the full-cloud baselines are trained with the same number of epochs, batch size, loss function and preprocessing as their counterpart hierarchical models.

Figure~\ref{fig:accuracy} shows the test accuracy of the hierarchically trained DNN when it is separated at different points along the architecture in comparison to the full-cloud training. For most of the separation points, the accuracy of the final exit of hierarchical training is close to the accuracy of the full-cloud training, even though the backpropagation is not connected between the edge and the cloud, and the information communicated in the forward pass is compressed. As a side benefit, the test accuracy of the early exit at the edge is shown. Although this accuracy is lower in comparison to the hierarchical cloud, it shows that our proposed method can also provide a level of robustness against network failures since the edge can independently handle the classification problem up to an acceptable accuracy. For example, at separation point 3 in the CIFAR-10 experiment, our proposed hierarchical training method provides an $89.29\%$ accuracy for the VGG-16 architecture in the final exit that is close to a $91.86\%$ accuracy in the baseline full-cloud training. In case of a communication failure, the early exit in our proposed hierarchical training method independently ensures an accuracy of $85.49\%$. In the CIFAR-10 experiments, the early exit accuracy becomes comparable to the final exit around the 4th separation point in the VGG-16 and ResNet-18 experiments. However, in the Tiny ImageNet experiments that are more complex problems, this does not happen until deeper separation points (8th separation point in VGG-16 and 6th separation point in ResNet-18). Although separating the DNN architectures in the deepest separation points (the gray areas in Figure~\ref{fig:accuracy}) is not practically desirable due to the high memory pressure at the edge, we keep them in the plots for completeness. This point is explained later.

Figure \ref{fig:memory} shows the number of parameters of the DNN at the edge and the cloud environments. These values are obtained by counting the parameters using both the hierarchical and the full-cloud training methods when the DNN is implemented. Of significance in these figures is that, for many of the separation points, especially the ones in the earlier layers, the number of parameters at the edge is significantly lower in comparison to the parameter count at the full cloud. As an example, at the separation point 3 in the CIFAR-10 experiment, there are just $1.73 \times 10^5$ parameters implemented at the edge in our proposed hierarchical training method for VGG-16 architecture while this value is equal to $1.53\times10^7$ for the full-cloud training. This result is desirable in edge-cloud frameworks since the memory of the edge is often confined. 

The number of parameters is close in the CIFAR-10 experiments and the Tiny ImageNet experiments as the architectures are similar. The main source of difference is the higher number of neurons in the last fully connected layers of the DNNs implemented in the Tiny ImageNet experiments due to the higher number of classes. This results in a slightly higher parameter count compared to the CIFAR-10 experiments. The gray areas on the right side of Figure \ref{fig:memory} show the deepest separation points where the number of parameters in the edge is more than the cloud. These separation points are not practically desirable due to the high memory pressure at the edge and are shown in this figure and the next figures for completeness (the gray areas in Figures \ref{fig:accuracy}-\ref{fig:communication}).

%%%%%%

Figure \ref{fig:compute} shows the computational burden of hierarchical training of DNNs at the edge and the cloud in comparison to the full-cloud training in terms of multiplications and accumulations (MACC). The value is measured by counting all the MACC operations in each of these models, on each device. The crucial point is that the computational burden of the edge is significantly low for many of the separation points, especially in the earlier ones. For example, there are only $1.90\times10^8$ MACC operations at the edge in the VGG-16 architecture in our proposed hierarchical training method for the CIFAR-10 experiment, while there are $9.58\times10^8$ MACC operations in the baseline full-cloud training. This is favorable for the edge-cloud systems since there are often restrictions in the computational resources of the edge devices which may cause high latency. As a side benefit, one can see that in all separation points, the computational burden of the cloud is also lower in comparison to the full-cloud training, which provides a lower cost of cloud services. The number of MACC operations is higher in Tiny ImageNet experiments in comparison to CIFAR-10 experiments mainly due to larger sizes of inputs in Tiny ImageNet dataset. It is worth mentioning that, as the ResNet-18 architecture that is used for CIFAR-10 dataset has a different initial convolution layer in comparison to the one that is used for Tiny ImageNet dataset, the size of the feature maps in both entire models is similar. (This initial convolution has a kernel size=$3$, a stride=$1$ and a padding=$1$ for the CIFAR-10 experiments while it has a kernel size=$7$, a stride=$2$ and a padding=$3$ for the Tiny ImageNet experiments.) As a result, the number of MACC operations is close for the CIFAR-10 experiment and the Tiny ImageNet experiment on ResNet-18. However, in the VGG-16 architecture the initial convolution layer is the same for the CIFAR-10 and the Tiny ImageNet experiments. This results in a significant difference in the number of MACC operations due to larger feature maps of the latter experiment.

Figure \ref{fig:communication} shows the communication requirements of the proposed hierarchical training method in comparison to the full-cloud training. This value is the measured size of the quantized feature maps and the scales that are communicated in the hierarchical training model and the raw images that are communicated in the full-cloud model. As shown in the plots, for all the separation points, the communication burden is lower in comparison to the full-cloud training as this method does not communicate in the backward pass and compresses the information that is sent to the cloud in the forward pass. For example, at separation point 3, $16384$ bits are communicated during the proposed hierarchical training method in VGG-16 architecture for the CIFAR-10 experiment, while in the baseline full-cloud training, $24576$ bits are communicated.
In this experiment using our proposed hierarchical training method, we choose the number of channels of the compression convolution in such a way that the size of the output feature map stays less than a predefined fixed value, that results in almost flat curves. The drop in the communication burden in the deepest separation points is due to the reduction of the original feature map size in the deep layers of the VGG-16 and ResNet-18 architectures. The Tiny ImageNet experiments have higher communication requirements in comparison to the CIFAR-10 experiments due to larger sizes of the inputs and the feature maps.

\begin{figure*}
\centering
\includegraphics[scale=0.4]{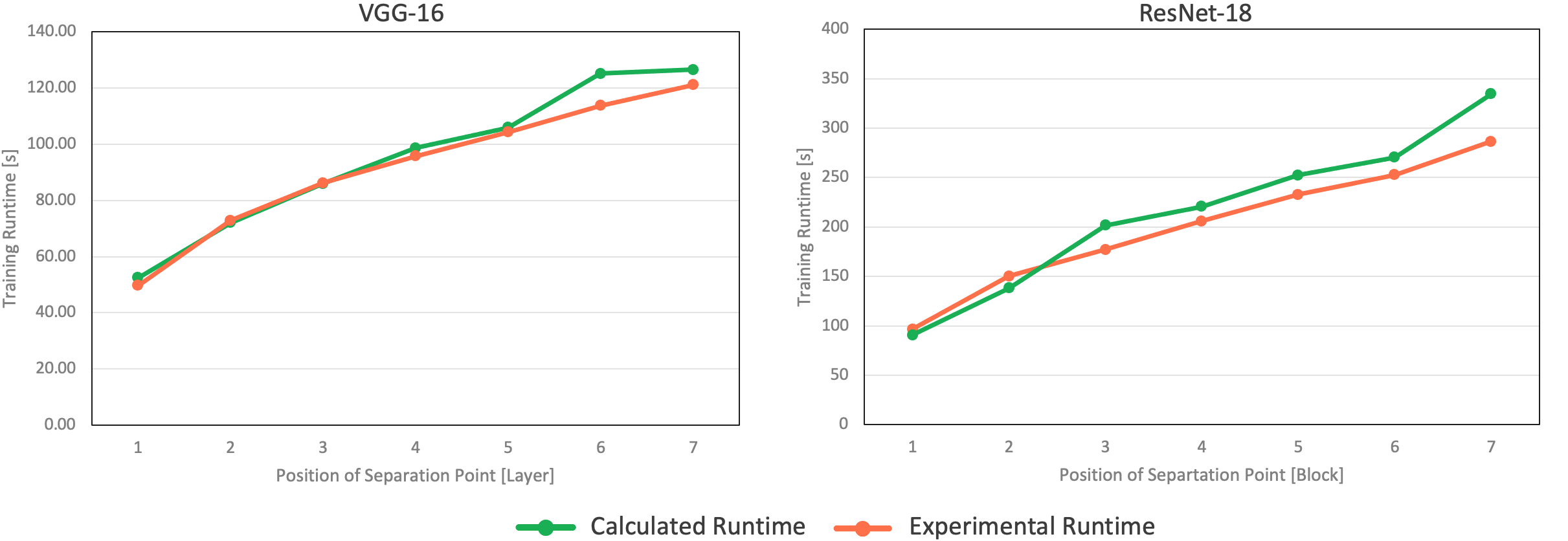}
\caption{The calculated runtime of one epoch of training in different separation points along the architecture in comparison to the experimental values, for the proposed hierarchical training framework implemented on a simplified edge-cloud system in the CIFAR-10 classification experiment. The left figure shows the results for VGG-16 and the right figure shows the results for ResNet-18. Our calculation method can provide a good estimation of the experimental values for most of the points.}
\label{fig:calc_exp_runtime}
\end{figure*}

The next key performance indicator is the training runtime. As improving the runtime of training is a main goal of our proposed hierarchical training method, we should have a proper estimation of this value to know where to separate a DNN architecture between the edge and the cloud (see Section~\ref{section:HierTrainEarly}). In Section~\ref{section:runtime_analysis}, we proposed a set of equations to estimate the runtime of training. In this part, we evaluate their performance by experimenting on a system of two devices, made of NVIDIA Quadro K620~\cite{nvidiak620} (edge) and NVIDIA GeForce RTX 2080 Ti~\cite{nvidia2080} (cloud). As mentioned before, we perform a simple forward pass runtime measurement, then we use Equations~(\ref{eq:tot_hier})-(\ref{eq:back_cloud}) to estimate the training runtime. We perform this for different separation points and compare the estimated runtime with the experimental runtime on the two devices. The devices are directly connected in this simplified experiment; consequently, the communication term of Equation~(\ref{eq:tot_hier}) is negligible. 

Figure~\ref{fig:calc_exp_runtime} shows the result of this experiment. It shows that our calculation method can provide a good estimation of runtime for different division points along the DNN architectures, and it can be used for selection of the separation point (see Section~\ref{section:separation_selection}). It is worth mentioning that for the VGG-16 experiment, the selected edge device memory could not handle the separation points after the 7th point; as a result, they do not exist in Figure~\ref{fig:calc_exp_runtime}.

\begin{figure*}[t]
\centering
\includegraphics[scale=0.48]{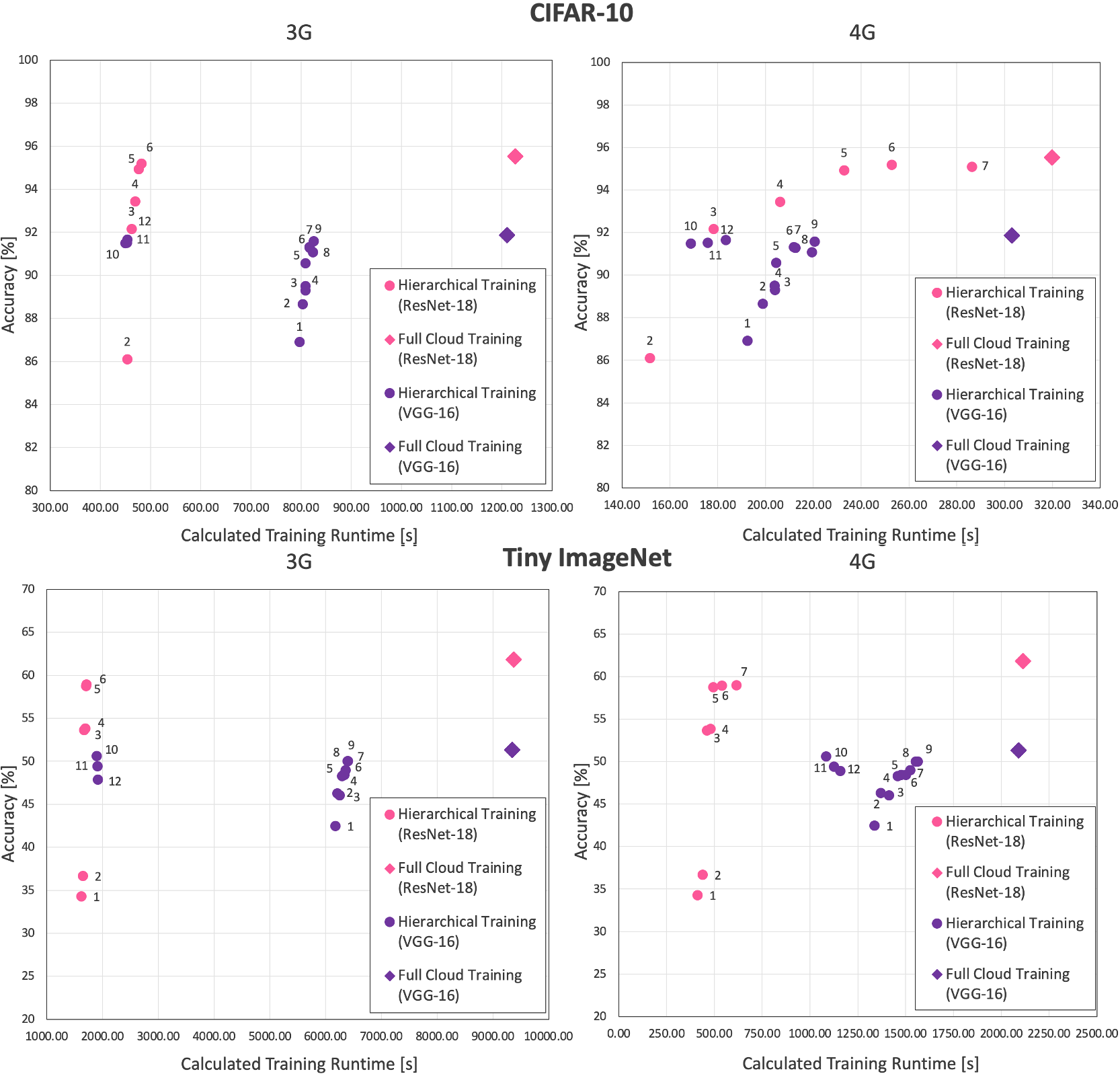}
\caption{The accuracy and the calculated runtime of the proposed hierarchical training method when it is separated at different points along the architecture in comparison to the full-cloud training in CIFAR-10 and Tiny ImageNet experiments. The label numbers show the positions of the separation points. The left figures show the results using 3G communication protocol and the right figures show the results using 4G communication protocol.}
\label{fig:hier_results_3G_4G}
\end{figure*}

%%%%%%%%%%%%%%%%%%%%

Using Equations (\ref{eq:tot_hier})-(\ref{eq:tot_full}) from Section~\ref{section:runtime_analysis} and the accuracies that have been measured in Figure \ref{fig:accuracy}, we obtain Figure~\ref{fig:hier_results_3G_4G}. This figure shows the accuracy versus the calculated runtime for one epoch of training for the different separation points in the DNNs for our proposed hierarchical training method in comparison to the full-cloud training baseline. The separation points are shown by the labels in the plot. These runtime values are computed for 3G and 4G communication protocols between the edge and the cloud. The selection of 3G and 4G networks is justified by their extensive global coverage, encompassing $94.9\%$ and $85.0\%$ of the world population, respectively, according to~\cite{ericsson}.  

The figures show that for both VGG-16 and ResNet-18 architectures used with CIFAR-10 and Tiny ImageNet datasets with 3G and 4G communication protocols, our proposed method can provide a good reduction in the runtime, while the penalty on accuracy is low for many of the separation points. The gain in runtime is higher when 3G communication protocol is used since the communication bandwidth is lower and reducing its burden can have a more significant effect on the total runtime. These figures help to choose the best hierarchical division between the edge and the cloud based on the specific accuracy and runtime requirements of the user. Consider that Figures \ref{fig:memory} and \ref{fig:compute} also help the user to satisfy the memory footprint and the computational cost requirements.

\vspace{5pt}
{\subsection{On-Device Experiments}\label{section:ondevice}}

In this section, we demonstrate the performance of the proposed hierarchical training method with on-device experiments. In this setting, the edge device has a low-resource NVIDIA Quadro K620~\cite{nvidiak620} chip with $863.2$ GFLOPS theoretical performance and the cloud has a high-end NVIDIA GeForce RTX 2080 Ti~\cite{nvidia2080} chip with $13.45$ TFLOPS theoretical performance. Appendix~\ref{app1} gives more details about the effect of changing the edge device computational resources on the proposed hierarchical training method. For the data communication stage, we simulate two different telecommunication protocols, 3G and 4G. 
Notice that in these setups, the data transfer bottleneck is the upload link which has an average speed of $1.1$ Mbps for 3G and $5.85$ Mbps for 4G in the United States~\cite{eshratifar_jointdnn_2020}. 

Table~\ref{table:runtime_cifar10} shows the results of our experiments for VGG-16 and ResNet-18 on CIFAR-10 dataset. The computational runtime results for one epoch of training and the final accuracies after training for all epochs are shown for the sake of comparison. In these experiments, the separation point 3 has been selected for both architectures as it provides a good balance between the accuracy, the edge memory, the computational cost and the runtime (see Figures \ref{fig:accuracy}-\ref{fig:hier_results_3G_4G}). Table~\ref{table:runtime_cifar10} shows that for the two different DNNs and the two telecommunication technologies, 
the hierarchical training scenarios provide a lower runtime in comparison to the full-cloud systems whilst having a marginal reduction in accuracy. In 3G communication, it results in $28.96\%$ improvement for VGG-16 and $60.91\%$ for ResNet-18 in terms of runtime. For 4G communication, the improvements are $13.78\%$ for VGG-16 and $36.26\%$ for ResNet-18.

Table~\ref{table:runtime_tinyimagenet} presents the results of on-device experiments using VGG-16 and ResNet-18 architectures on Tiny ImageNet dataset. In these experiments, separation points 5 and 6 have been selected for VGG-16 and ResNet-18 architectures respectively. That provides a proper balance between the accuracy, the edge memory, the computational cost and the runtime (see Figures \ref{fig:accuracy}-\ref{fig:hier_results_3G_4G}). Similarly to CIFAR-10, for both DNN architectures and telecommunication technologies, 
the proposed hierarchical training method provides a lower runtime in comparison to the full-cloud system whilst having a marginal reduction in accuracy. In 3G communication, it results in a $25.14\%$ improvement for VGG-16 and a $81.25\%$ improvement for ResNet-18 in terms of runtime. For 4G communication, the improvements are $4.72\%$ for VGG-16 and $73.97\%$ for ResNet-18. The improvement in ResNet-18 is more significant since it has a specific initial convolution for datasets with large image samples (like Tiny ImageNet) that keeps the intermediate activations small as we discussed in Section~\ref{section:perf_analysis}.

As expected, the proposed hierarchical method is more advantageous for less efficient communication links since typically, the communication link is the bottleneck of the overall efficiency of the hierarchical systems. Moreover, it is shown that our calculations generally provide a good estimation in comparison to the experiments. 
The calculated values are slightly different compared to the experimental ones, due to the complexities that exist in the real devices and experiments that are not counted in our simplified calculations, like the overhead for the communication between different parts inside a single GPU.

\begin{table*}[t]
  
  \centering
  \footnotesize
  \caption{Runtime comparison for CIFAR-10 experiments across different neural networks trained using different deep training strategies and telecommunication technologies
  } 
  \begin{tabular}{|c|c|c|cc|cc|}
  
    \hline
        \multirow{2}{*}{Neural Network} & \multirow{2}{*}{Training Strategy} & \multirow{2}{*}{Accuracy[\%]} & \multicolumn{2}{c|}{3G Runtime[s]} & \multicolumn{2}{c|}{4G Runtime[s]} \\
    %\cline{3-6} 
    & &  &Experiment & Calculation & Experiment & Calculation \\
    \hline\hline

    VGG-16   & Hierarchical  & 89.15 & 840.60 &  808.65  & 233.88  &  203.96 \\
    \cline{2-2}
    %\cline{2-2}
     (Divided at Point 3) & Full-cloud  & 91.86 & 1183.28 & 1210.05 & 271.25  & 303.02 \\
    \hline\hline

    ResNet-18  & Hierarchical & 92.21 & 466.17 & 462.25  &   180.08   &  178.31  \\
    \cline{2-2}
     (Divided at Point 3) & Full-cloud & 95.52 & 1192.51  & 1226.97 &   282.52    & 319.93 \\
     \hline
     
\end{tabular}

 \label{table:runtime_cifar10}
\end{table*}

\vspace{5pt}
\begin{table*}[t]
  
  \centering
  \footnotesize
  \caption{Runtime comparison for Tiny ImageNet experiments across different neural networks trained using different deep training strategies and telecommunication technologies
  } 
  \begin{tabular}{|c|c|c|cc|cc|}
  
    \hline
        \multirow{2}{*}{Neural Network} & \multirow{2}{*}{Training Strategy} & \multirow{2}{*}{Accuracy[\%]} & \multicolumn{2}{c|}{3G Runtime[s]} & \multicolumn{2}{c|}{4G Runtime[s]} \\
    %\cline{3-6} 
    & &  &Experiment & Calculation & Experiment & Calculation \\
    \hline\hline

    VGG-16   & Hierarchical  & 48.42 & 6849.31  &    6361.02 & 2014.92  & 1523.47  \\
    \cline{2-2}
    %\cline{2-2}
     (Divided at Point 6) & Full-cloud  & 51.33 & 9149.34 &  9346.74  &  2114.65 & 2090.43 \\
    \hline\hline

    ResNet-18  & Hierarchical & 58.76 & 1718.33 &  1701.55 &  493.93    &  495.29  \\
    \cline{2-2}
     (Divided at Point 5) & Full-cloud & 61.83 &  9164.88 & 9370.47 &  1897.39     & 2114.15 \\
     \hline
     
\end{tabular}

 \label{table:runtime_tinyimagenet}
\end{table*}

\vspace{5pt}

\begin{table*}[t]
  
  \centering
  \footnotesize
  \caption{Ablation study to evaluate the effect of each component of the proposed method on CIFAR-10 dataset
  } 
  \begin{tabular}{|c|c|c|r|r|r|} \hline 
 \multirow{2}{*}{Neural Network}& \multirow{2}{*}{Model}& \multirow{2}{*}{Accuracy [\%]}&  \multirow{2}{*}{3G Runtime[s]}&  \multirow{2}{*}{4G Runtime[s]}& Communication  \ \   \\
   & & &  &  &  Requirements [bits]\\
    \hline\hline
    
    & \multirow{2}{*}{Full-Cloud Model} & \multirow{2}{*}{91.86} & \multirow{2}{*}{1183.28  } &  \multirow{2}{*}{271.25   } & \multirow{2}{*}{24576   }  \\ 
    & &  &  &    &   \\ \Cline{0.8pt}{2-6}
       & Hierarchical Model without & \multirow{2}{*}{91.86} & \multirow{2}{*}{95593.78   } &  \multirow{2}{*}{18143.55   } & \multirow{2}{*}{2097152   }  \\ 
    &Early Exit \& Quantization&  &  &    &   \\ \cline{2-6}
    VGG-16 & Hierarchical Model with  & \multirow{2}{*}{84.88} & \multirow{2}{*}{53799.57  } &  \multirow{2}{*}{10182.331   } & \multirow{2}{*}{1179648   }  \\ 
     (Divided at Point 3) & Quantization \& without Early Exit &  &  &  &  \\ \cline{2-6}
 & Hierarchical Model with & \multirow{2}{*}{88.23} & \multirow{2}{*}{6070.03   }& \multirow{2}{*}{1224.32   } & \multirow{2}{*}{131072   } \\
    
    & Early Exit \& without Quantization & & & & \\\Cline{0.8pt}{2-6}

& Proposed Hierarchical Model & \multirow{2}{*}{89.15} & \multirow{2}{*}{840.60   }& \multirow{2}{*}{233.88   } & \multirow{2}{*}{16384   } \\
    
    & with Early Exit \& Quantization & & & & \\
    \hline\hline

       & \multirow{2}{*}{Full-Cloud Model} & \multirow{2}{*}{95.52} & \multirow{2}{*}{1192.51   } &  \multirow{2}{*}{282.52   } & \multirow{2}{*}{24576   }  \\ 
    & &  &  &    &   \\ \Cline{0.8pt}{2-6}
       & Hierarchical Model without & \multirow{2}{*}{95.52} & \multirow{2}{*}{95621.85   } &   \multirow{2}{*}{18170.84   } &  \multirow{2}{*}{2097152   } \\ 
    &Early Exit \& Quantization&  &  &    &   \\ \cline{2-6}
    ResNet-18 & Hierarchical Model with & \multirow{2}{*}{88.40} & \multirow{2}{*}{53803.76   } & \multirow{2}{*}{10187.41   } &  \multirow{2}{*}{1179648   } \\ 
     (Divided at Point 3) & Quantization \& without Early Exit &  &  &  & \\ \cline{2-6}
 & Hierarchical Model with & \multirow{2}{*}{91.81} & \multirow{2}{*}{3077.81   } & \multirow{2}{*}{651.74   } & \multirow{2}{*}{65536   } \\
    
    & Early Exit \& without Quantization & & & &\\
    \Cline{0.8pt}{2-6}
    & Proposed Hierarchical Model & \multirow{2}{*}{92.21} & \multirow{2}{*}{466.17   }& \multirow{2}{*}{180.08   } & \multirow{2}{*}{8192   } \\
    
    & with Early Exit \& Quantization & & & & \\
    \hline
     
\end{tabular}

 \label{table:ablation}
\end{table*}
{\section{Ablation Study}\label{section:ablation_study}}
 
In this section, an ablation study is performed to demonstrate the effectiveness of the different blocks of the proposed method. Table~\ref{table:ablation} exhibits the different ablation studies that are done based on the proposed hierarchical training method. To facilitate the comparison, the full-cloud model and the proposed hierarchical model are respectively positioned in the first and the last rows within each neural network architecture. The first ablated model is a hierarchical model without the early exit and the quantization. This model communicates the full-precision feature map at the separation point between the edge and the cloud in the forward pass and the full-precision gradients in the backward pass. Consequently, the runtime and the communication requirements of this model are significantly higher than the ones of the proposed hierarchical model and even the full-cloud model. The accuracy of this model is similar to the full-cloud model as it does not contain any modifications in terms of the DNN architecture. The second ablated model contains a quantization similar to the proposed hierarchical method but does not have an early exit. Since quantization is not a differentiable layer, we used Straight Through Estimator (STE)~\cite{ bengio2013estimating, zhou2016dorefa} to estimate the backward pass at the separation point. As we see in the results, since the backward pass is estimated by STE to update the edge parameters, the accuracy of this model is lower than the proposed hierarchical method and the full-cloud method. Additionally, as the backward pass is still communicated in this model, the communication requirements and the runtime are also higher than the proposed method. The third ablated model is a hierarchical model with the early exit and without the quantization. In this model, as the backward pass is not communicated due to the help of the early exit, the communication requirements and the runtime are lower than the two other ablated models. However, the communication requirements and the runtime are still worse than the proposed hierarchical model since the communicated feature map in the forward pass is not quantized. The accuracy of this ablated model is close to the proposed hierarchical method but slightly lower. This can be a result of the quantization layer acting as a regularizer during the training phase by adding noise to the model in the forward pass, while not negatively impacting the backward pass due to the existence of the early exit~\cite{nagel2021white}.
\vspace{5pt}

\begin{figure*}[h]
\centering
\includegraphics[scale=0.41]{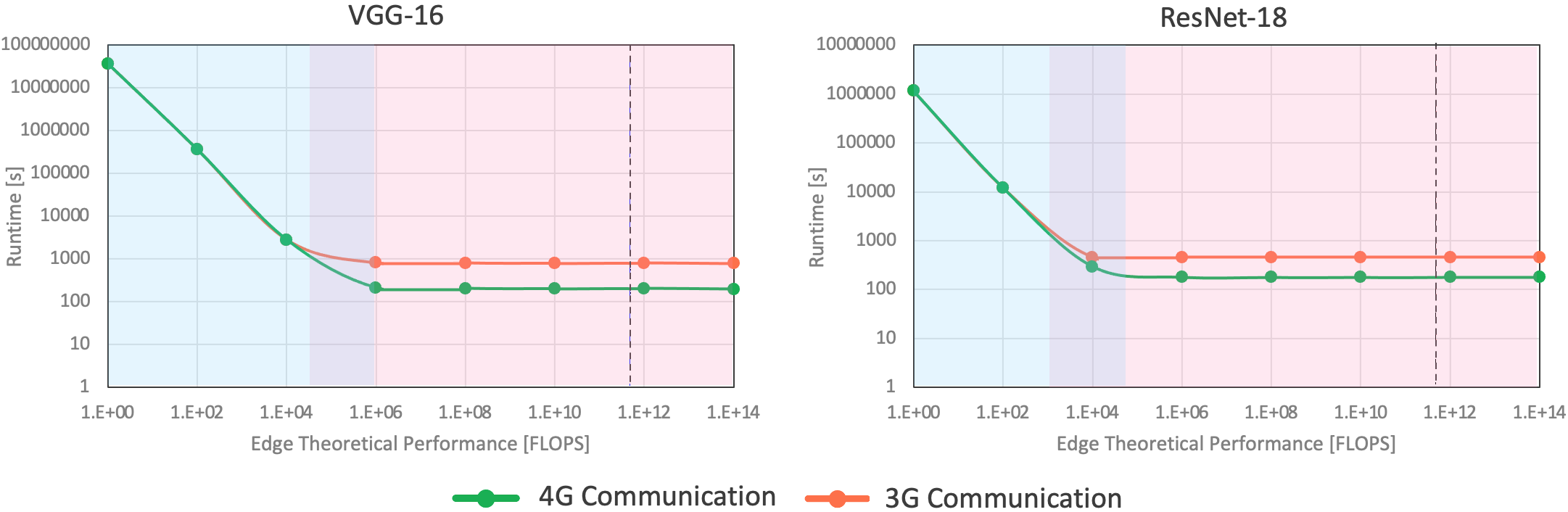}
\caption{The required runtime of training for $1$ epoch of a DNN architecture using the proposed hierarchical training method vs. the performance of the edge device. The left figure shows the results for VGG-16 architecture and the right figure shows the results for ResNet-18. The values are calculated using Equations~(\ref{eq:tot_hier})-(\ref{eq:tot_full}) while keeping the cloud device performance the same ($13.45$ TFLOPS). The dashed lines show the edge device that we used for the on-device experiments. The blue area highlights where the edge backward pass delay is dominant in the total runtime and the red area highlights where the sum of communication, cloud forward pass and cloud backward pass delays are dominant. The axes are shown on a logarithmic scale.}
\label{fig:edge_computational_resource}
\end{figure*}

{\section{Summary and Future Works}\label{section:conclusion}}

In this study, a novel hierarchical training approach for DNN architectures in edge-cloud scenarios has been proposed. It provides less communication cost, lower runtime, higher privacy for the user and improved robustness against possible network failures compared to the full-cloud training. We performed simulations on different neural network architectures and implemented the proposed approach on a two-device framework and validated its superiority with respect to the full-cloud training on different datasets.

In the domain of hierarchical training of neural networks, a further topic to investigate is the design of hierarchical-friendly neural network architectures.
 As we saw in this study, the available DNN architectures are not made to be executed hierarchically. They have issues like the larger size of intermediate feature maps in comparison to the inputs, which reduces the effectiveness of hierarchical training approaches. New neural network architectures could be developed that inherently consider this issue. They could also take into account other important points such as keeping a low computational cost in the parts that are executed at the edge. 
 \appendices
 \vspace{2pt}
{\section{Edge Resource Effect Evaluation}\label{app1}}

In this appendix, we evaluate the effect of changing the edge resources by calculating the runtime of our proposed hierarchical method. We use Equations~(\ref{eq:tot_hier})-(\ref{eq:tot_full}) to calculate the runtime for a range of performances of the edge. The calculations are done for the CIFAR-10 classification experiment on VGG-16 and ResNet-18 architectures. We kept the cloud device similar to our previous experiments (NVIDIA GeForce RTX 2080 Ti~\cite{nvidia2080} with $13.45$ TFLOPS theoretical performance).  

Figure~\ref{fig:edge_computational_resource} demonstrates the results. For both VGG-16 and ResNet-18 architectures, when the edge performance is higher than a specific point, the curves become almost flat and they do not depend significantly on edge performance anymore. The reason for this is that in the proposed hierarchical training method, the backward pass of the edge is done in parallel with the edge-cloud communication, the forward pass and the backward pass of the cloud. If the edge is fast enough, the latter becomes dominant in terms of the total runtime (the red areas in Figure~\ref{fig:edge_computational_resource}). Consequently, as the total runtime does not depend on the edge performance anymore, the curve becomes flat. However, when the edge performance is low, the backward pass of the edge becomes dominant in the total runtime and the total runtime becomes dependent on the edge performance (the blue areas in Figure~\ref{fig:edge_computational_resource})). In this case, the 3G communication and 4G communication curves overlap since the edge-cloud communication is not determinant in the total runtime anymore.

\vspace{5pt}
\bibliographystyle{unsrt}

\end{document}